\documentclass[conference]{IEEEtran}
\IEEEoverridecommandlockouts
\usepackage{cite}
\usepackage{amsmath,amssymb,amsfonts}
\usepackage{algorithmic}
\usepackage{graphicx}
\usepackage{textcomp}

\usepackage{subcaption}
\usepackage{algorithm}

\usepackage{xcolor}
\def\BibTeX{{\rm B\kern-.05em{\sc i\kern-.025em b}\kern-.08em
    T\kern-.1667em\lower.7ex\hbox{E}\kern-.125emX}}

\makeatletter
\newcommand{\linebreakand}{%
  \end{@IEEEauthorhalign}
  \hfill\mbox{}\par
  \mbox{}\hfill\begin{@IEEEauthorhalign}
}
\makeatother

\begin{document}

\title{Efficient Geothermal Well-Control Optimization via Diffusion-Surrogate Reinforcement Learning\\
}

\author{\IEEEauthorblockN{Ruimin Dai}
\IEEEauthorblockA{\textit{School of Computing} \\
\textit{Clemson University}\\
Clemson, SC, USA \\
ruimind@clemson.edu}
\and
\IEEEauthorblockN{Guodong Chen}
\IEEEauthorblockA{\textit{Energy Geoscience Division} \\
\textit{Lawrence Berkeley National Laboratory}\\
\textit{University of California Berkeley}\\
Berkeley, CA, USA \\
gchen6@lbl.gov}

\and
\IEEEauthorblockN{Randy Harsuko}
\IEEEauthorblockA{\textit{Energy Geoscience Division} \\
\textit{Lawrence Berkeley National Laboratory}\\
\textit{University of California Berkeley}\\
Berkeley, CA, USA \\
mharsuko@lbl.gov}

\linebreakand

\and
\IEEEauthorblockN{Kunpeng Liu}
\IEEEauthorblockA{\textit{School of Computing} \\
\textit{Clemson University}\\
Clemson, SC, USA \\
kunpenl@clemson.edu}

\and
\IEEEauthorblockN{Nori Nakata\textsuperscript{*}}
\IEEEauthorblockA{\textit{Energy Geoscience Division} \\
\textit{Lawrence Berkeley National Laboratory}\\
Berkeley, CA, USA \\
nnakata@lbl.gov}
}



\maketitle

\begingroup
\renewcommand{\thefootnote}{*}
\footnotetext{Corresponding author.}
\endgroup

\begin{abstract}

Real-time decision-making for enhanced geothermal systems (EGS) is challenging because long-term production periods involve high-dimensional control spaces and a large number of time-consuming high-fidelity hydrothermal simulations. Reinforcement learning provides a natural framework for state-dependent sequential control, but direct policy training with numerical simulators is computationally expensive. To address this issue, we propose a diffusion-surrogate guided reinforcement learning framework for long-horizon EGS well-control optimization. The reservoir temperature and pressure fields are used as system states, while injection rates are selected as control actions. A learned surrogate environment is constructed using conditional diffusion models to predict the evolution of reservoir temperature and pressure fields and a separate reward model to estimate the corresponding economic return. The surrogate environment is then integrated with Proximal Policy Optimization (PPO) for efficient policy training. Experiments on a fractured EGS benchmark show that the diffusion surrogate can accurately reproduce reservoir-state evolution over multiple control stages. The resulting surrogate-assisted PPO policy achieves competitive well-control performance compared with direct simulator-based PPO and existing optimization methods, while substantially reducing the dependence on expensive high-fidelity simulations. These results demonstrate the potential of diffusion-based surrogate environments for efficient reinforcement learning in geothermal well-control optimization.

\end{abstract}


\section{Introduction}

\begin{figure}[t]
    \centering
    \includegraphics[width=0.95\linewidth]{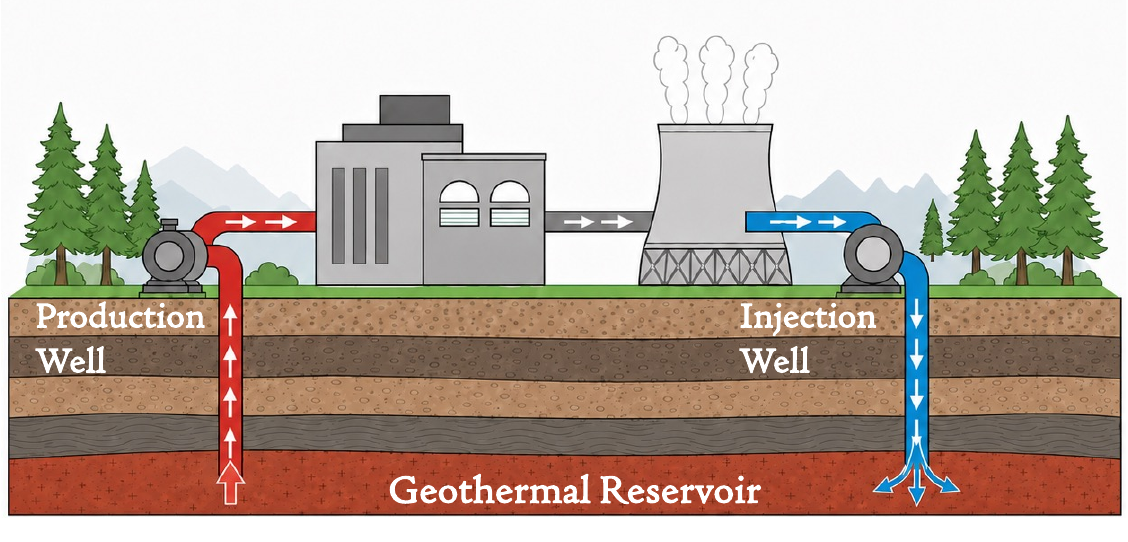}
    \caption{Schematic illustration of an enhanced geothermal system (EGS). Cold fluid is injected into the geothermal reservoir through an injection well, heated as it circulates through the subsurface formation, and recovered through a production well for surface energy utilization.}
    \label{fig:EGS}
\end{figure}

Enhanced Geothermal Systems (EGS) offer a practical approach to extracting geothermal energy from high-temperature rock formations where natural permeability is insufficient for conventional geothermal production~\cite{parisio2020modeling,olasolo2016enhanced,liu2026beyond}. As illustrated in Fig.~\ref{fig:EGS}, relatively cold fluid is injected into the geothermal reservoir through injection wells, circulates through fractures and the surrounding rock matrix, and absorbs heat from the hot formation. The heated fluid is then recovered through production wells and transported to the surface for electricity generation or direct heat use. This circulation process enables continuous heat extraction from the subsurface reservoir. Because fractures serve as the dominant pathways for fluid transport, EGS performance is strongly controlled by the coupled behavior of subsurface flow and heat transfer~\cite{karvounis2022discrete,yao2018numerical,praditia2018multiscale}. Changes in well operation can redistribute reservoir pressure and redirect fluid circulation, which in turn alters the evolution of the temperature field. These effects directly influence how effectively the reservoir volume is swept, when cold-water breakthrough occurs at production wells, and how much thermal energy can be recovered over long-term operation.

Properly designed well operating strategies are therefore critical to the long-term economic performance of EGS projects. Well-control optimization seeks to determine suitable control settings at different stages of geothermal production so that heat can be extracted efficiently while maintaining economic viability over the project lifetime~\cite{wang2023deep,chen2024surrogate,juliusson2013optimization}. Depending on the operating scheme, the decision variables may include injection rates, production rates, and bottomhole pressures. These decisions affect not only the immediate production response but also the subsequent evolution of reservoir pressure, temperature, and overall production potential.

For this reason, well-control optimization cannot be reduced to maximizing short-term injection rates, production rates, or thermal output at individual time steps~\cite{juliusson2013optimization,pollack2019accounting,chen2015efficient}. Control decisions must be evaluated over the full development horizon, where increased heat production is weighed against costs associated with fluid injection, production, and other operational activities. Net present value (NPV) is commonly used to capture this trade-off in a single economic objective. From this perspective, EGS well-control optimization can be formulated as a sequential decision-making problem that searches for a time-dependent control policy capable of maximizing cumulative economic returns over the entire production period.

Reinforcement learning (RL) is well suited to long-horizon sequential decision problems in which current control actions affect future system states and returns~\cite{kaelbling1996reinforcement,sutton1999policy,mnih2015human,schulman2015trust}. Rather than optimizing a fixed sequence of predetermined control variables directly, RL learns a policy through interactions between an agent and its environment, with actions selected from the current state to maximize cumulative reward over the decision horizon. This formulation has been used in engineering control problems including oil and gas reservoir production optimization, where reservoir states are provided as observations, well-control variables define the action space, and net present value (NPV) is used as the reward~\cite{zhang2022training,nasir2023deep}. The same sequential structure arises in EGS well-control optimization. Injection decisions at each control stage modify the reservoir pressure and temperature fields, and these changes affect heat extraction and economic performance in later stages. RL can therefore condition well-control decisions on the evolving reservoir state and optimize them with respect to long-term economic return rather than the immediate response at a single control step.

Although optimization methods have been applied to well-control optimization, three key limitations remain for long-term EGS well-control optimization:

\begin{enumerate}
    \item \textbf{High-dimensional well-control spaces increase optimization difficulty.} Optimizing multiple wells over multiple control stages produces a large number of decision variables. Combined with the nonlinear relationship between well controls and reservoir response, this high-dimensional search space can require many evaluations to converge and may prevent the optimizer from identifying high-quality control schemes within a limited computational budget~\cite{wang2019well,salehian2022robust}.

    \item \textbf{Hydrothermal simulations create a substantial computational bottleneck.} Each candidate control scheme must typically be evaluated using a computationally expensive hydrothermal simulation~\cite{xue2026robust,chen2024surrogate,sasso2023posterior}. Because optimization requires repeated evaluations of different control schemes, simulation cost can dominate the overall optimization process. The same issue becomes more pronounced in RL, where policy learning requires a large number of interactions with the simulation environment.

    \item \textbf{Predefined control sequences do not explicitly account for evolving reservoir states.} Most existing optimization methods determine the well-control sequence for the full production horizon in advance. These open-loop strategies do not establish an explicit mapping from the evolving reservoir state to subsequent control actions. Changes in reservoir temperature and pressure during production therefore cannot be directly incorporated into later well-control decisions, limiting adaptation to the reservoir conditions encountered during operation~\cite{jansen2008model}.
\end{enumerate}

To address these limitations, we propose a diffusion-surrogate reinforcement learning framework for long-horizon EGS well-control optimization. We formulate well control as a sequential decision problem, where reservoir temperature and pressure fields represent the system state and injection rates are selected as control actions. To reduce the cost of repeated hydrothermal simulations, we construct a surrogate environment in which conditional diffusion models predict the next reservoir state from the current state and action, while a separate reward model estimates the corresponding NPV reward. The learned surrogate environment is then integrated with Proximal Policy Optimization (PPO)~\cite{schulman2017proximal} for efficient policy training, while the high-fidelity COMSOL simulator is used for final policy evaluation. Our contributions are summarized as follows:

\begin{itemize}
    \item \textbf{Problem.} We formulate EGS well-control optimization as a state-dependent sequential decision problem. Rather than optimizing a fixed open-loop control sequence over the full production horizon, the formulation learns a policy that maps the evolving reservoir temperature and pressure fields to well-control actions.

    \item \textbf{Algorithm.} We develop a diffusion-based surrogate environment for RL training. Conditional diffusion models approximate reservoir state transitions conditioned on the current state and injection action, and a separate reward surrogate estimates the resulting economic return. Integrating these models with PPO reduces the number of high-fidelity hydrothermal simulations required during policy learning.

    \item \textbf{Evaluation.} We evaluate the proposed framework on a fractured EGS well-control problem, examining surrogate prediction accuracy, control performance, and computational cost. The results show that the learned policy achieves competitive performance while requiring substantially fewer high-fidelity simulation evaluations.
\end{itemize}

\section{Problem Statement}

\begin{figure*}[htbp]
    \centering
    \includegraphics[width=0.95\textwidth]{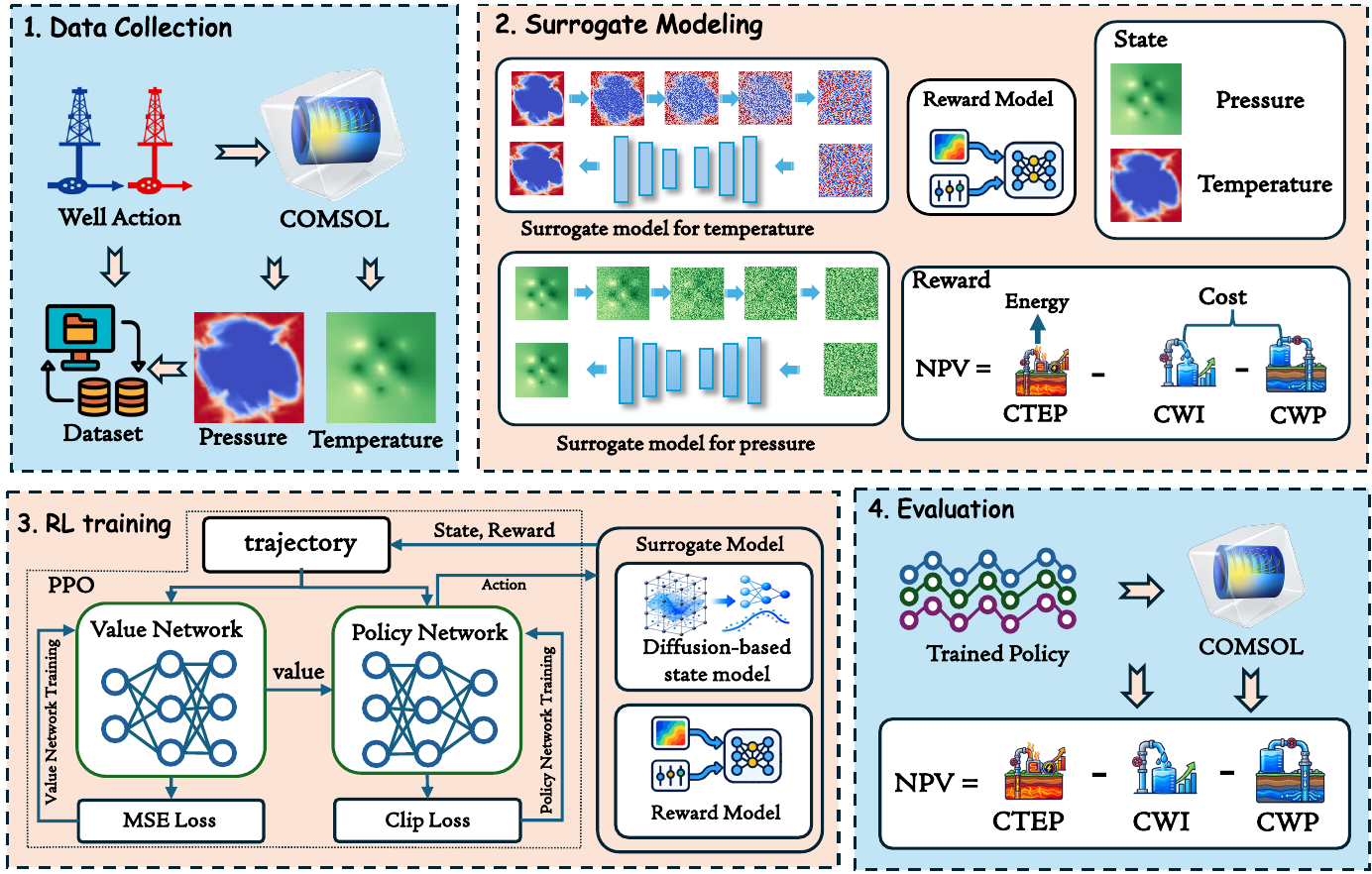}
    \caption{Overview of the proposed diffusion-surrogate reinforcement learning framework for EGS well-control optimization, including data collection, surrogate environment construction, PPO policy training, and high-fidelity policy evaluation.}
    \label{fig:framework}
\end{figure*}

\subsection{Hydrothermal Dynamics of Fractured EGS}
The enhanced geothermal system considered in this work consists of a low-permeability rock matrix intersected by highly permeable fractures. The fractures provide preferential pathways for fluid circulation, while the surrounding hot rock matrix serves as the primary source of thermal energy. During geothermal production, relatively cold water is injected into the reservoir through injection wells. The resulting pressure gradients drive fluid through the fracture network toward the production wells. As the injected fluid circulates through the reservoir, heat is transferred from the hot rock to the fluid and subsequently transported along the flow pathways. Therefore, the reservoir response is governed by the coupled evolution of fluid flow and heat transfer.

Fluid transport in the fractured reservoir follows mass conservation. For the porous medium, the governing equation can be expressed as
\begin{equation}
\frac{\partial (\rho_f \phi)}{\partial t}
+
\nabla \cdot (\rho_f \mathbf{u})
=
\Psi_{mf}+Q_m,
\end{equation}
where $\rho_f$ denotes the fluid density, $\phi$ is the porosity, $\mathbf{u}$ is the Darcy velocity, $\Psi_{mf}$ represents fluid exchange between the rock matrix and fractures, and $Q_m$ denotes the mass source or sink associated with well operation. The fluid velocity is governed by Darcy's law
\begin{equation}
\mathbf{u}
=
-\frac{k}{\mu}
\left(
\nabla p+\rho_f g\nabla D
\right),
\end{equation}
where $k$ is the permeability, $\mu$ is the fluid viscosity, $p$ is the reservoir pressure, $g$ denotes gravitational acceleration, and $D$ represents the elevation coordinate. Because the fracture permeability is substantially higher than the matrix permeability, the fracture network forms the dominant pathways for pressure-driven fluid transport.
The reservoir temperature evolution is governed by an energy balance that accounts for both advective and conductive heat transfer,
\begin{equation}
(\rho C)_{\mathrm{eff}}
\frac{\partial T}{\partial t}
+
\rho_f C_f \mathbf{u}\cdot\nabla T
-
\nabla\cdot
\left(
\lambda_{\mathrm{eff}}\nabla T
\right)
=
E_{mf}+Q_h,
\end{equation}
where $T$ denotes temperature, $(\rho C)_{\mathrm{eff}}$ is the effective volumetric heat capacity of the fluid--rock system, $C_f$ is the fluid heat capacity, $\lambda_{\mathrm{eff}}$ is the effective thermal conductivity, $E_{mf}$ represents heat exchange between the matrix and fractures, and $Q_h$ denotes the thermal source or sink.

These coupled hydrothermal processes establish a direct relationship between well operation and reservoir evolution. Changing the injection rates modifies the reservoir pressure gradients and fluid circulation pathways, which alters heat transport and the spatial distribution of reservoir temperature. Consequently, well-control decisions influence not only the instantaneous production response but also the future thermal and hydraulic states of the reservoir. In this work, the coupled hydrothermal equations are solved using the high-fidelity COMSOL simulator, and the resulting temperature and pressure fields are used to characterize the reservoir state.

\subsection{Reinforcement Learning Formulation}

Based on the coupled hydrothermal dynamics described above, we formulate EGS well-control optimization as a finite-horizon Markov decision process (MDP). At each control step $t$, the reservoir state is represented by
\begin{equation}
s_t=[T_t,P_t],
\end{equation}
where $T_t$ and $P_t$ denote the reservoir temperature and pressure fields, respectively. These two fields characterize the thermal and hydraulic conditions resulting from the coupled fluid-flow and heat-transfer processes.

Given the current reservoir state, a parameterized policy $\pi_{\theta}$ determines the well-control action
\begin{equation}
a_t=[q_{1,t},\ldots,q_{N_I,t}],
\end{equation}
where $q_{i,t}$ denotes the injection rate of the $i$-th injection well and $N_I$ is the number of injection wells. Applying $a_t$ changes the pressure distribution and fluid circulation within the fractured reservoir, leading to the next reservoir state $s_{t+1}$ according to the underlying hydrothermal dynamics.

The economic return at each control stage is evaluated using the net present value (NPV), which balances thermal energy production against the costs of fluid injection and production. The objective is to learn a state-dependent policy that maximizes the cumulative economic return over the entire production horizon:
\begin{equation}
\underset{\theta}{\operatorname{arg\,max}}\;
\mathbb{E}_{\pi_\theta}
\left[
\sum_{t=0}^{H-1}
\gamma^t
\left(
\mathrm{CTEP}_t r_e
-
\mathrm{CWI}_t r_i
-
\mathrm{CWP}_t r_p
\right)
\right],
\end{equation}
where $H$ denotes the number of control stages and $\gamma$ is the discount factor. $\mathrm{CTEP}_t$, $\mathrm{CWI}_t$, and $\mathrm{CWP}_t$ denote the thermal energy production, fluid injection, and fluid production during control stage $t$, respectively, while $r_e$, $r_i$, and $r_p$ represent the corresponding thermal energy price, fluid injection cost, and fluid production cost.

Through this formulation, the well-control problem is represented as a controlled dynamical process,
\begin{equation}
(s_t,a_t)\rightarrow s_{t+1},
\end{equation}
where the transition is governed by the coupled fractured-fluid-flow and heat-transfer dynamics. Rather than optimizing a predefined control sequence over the entire production horizon, the objective is to learn a feedback policy that continuously determines well-control actions according to the evolving reservoir temperature and pressure fields.

\section{Method}

\subsection{Method Overview}


As illustrated in Fig.~\ref{fig:framework}, the proposed framework consists of four parts: data collection, surrogate modeling, RL training, and evaluation. First, high-fidelity COMSOL simulations are used to generate reservoir-state transitions and corresponding economic returns under different well-control actions. These data are then used to construct a surrogate environment, where conditional diffusion models predict the evolution of reservoir temperature and pressure fields and a reward model estimates the corresponding NPV. PPO subsequently interacts with the learned surrogate environment to optimize a state-dependent well-control policy. Finally, the trained policy is evaluated using the original COMSOL simulator. The following sections focus on the two core algorithmic modules of the framework: surrogate environment modeling and PPO-based policy optimization.

\subsection{Diffusion-Based Surrogate Model}
\begin{figure}[h]
    \centering
    \includegraphics[width=0.95\linewidth]{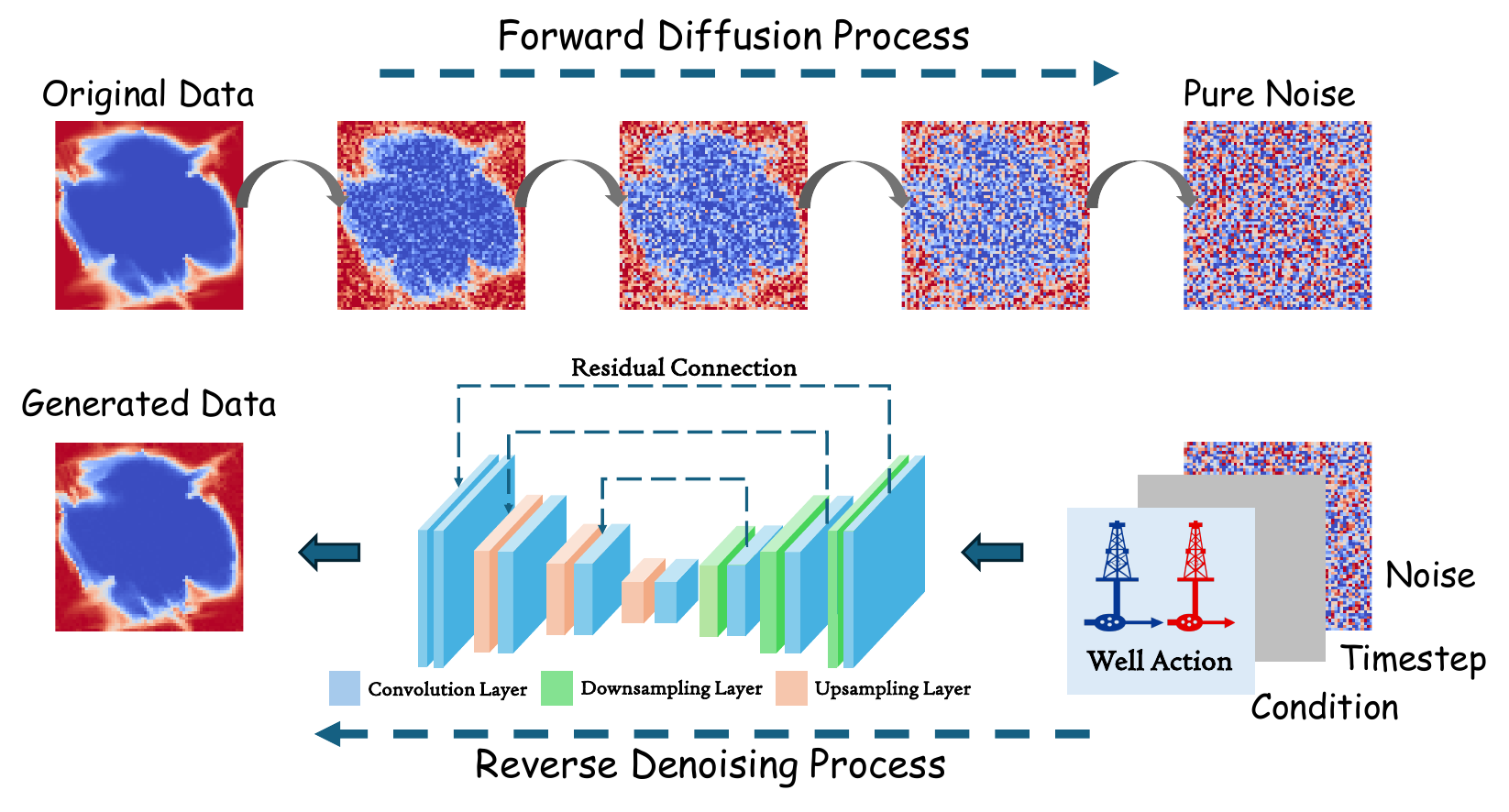}
    \caption{Conditional diffusion process for reservoir-state prediction}
    \label{fig:diffusion}
\end{figure}
To replace the high-fidelity simulator during policy optimization, the surrogate environment should approximate the reservoir response to each well-control action. Given the current state $\mathbf{s}_t$ and action $\mathbf{a}_t$, the environment is expected to provide the next reservoir state and the corresponding reward. We model these two components separately. The diffusion-based surrogate approximates the reservoir state transition
\begin{equation}
(\mathbf{s}_t,\mathbf{a}_t)
\rightarrow
\hat{\mathbf{s}}_{t+1},
\qquad
\mathbf{s}_t=[\mathbf{T}_t,\mathbf{P}_t],
\end{equation}
while the economic reward is estimated by a separate surrogate model introduced in the next subsection.

The reservoir state consists of high-dimensional temperature and pressure fields, whose evolution depends on both the current spatial distribution and the applied well-control action. We therefore formulate the state prediction using a conditional Denoising Diffusion Probabilistic Model (DDPM), which learns to generate the next reservoir field conditioned on $(\mathbf{s}_t,\mathbf{a}_t)$. Two independent diffusion models are trained for temperature and pressure prediction, respectively.

\paragraph{Conditional DDPM.}
For either temperature or pressure prediction, let $\mathbf{x}_0$ denote the target field at the next control stage. DDPM defines a forward diffusion process that gradually perturbs $\mathbf{x}_0$ with Gaussian noise over $K$ diffusion steps. The transition at diffusion step $k$ is
\begin{equation}
q(\mathbf{x}_k|\mathbf{x}_{k-1})
=
\mathcal{N}
\left(
\sqrt{1-\beta_k}\mathbf{x}_{k-1},
\beta_k\mathbf{I}
\right),
\end{equation}
where $\{\beta_k\}_{k=1}^{K}$ is a predefined noise schedule. Let
$\alpha_k=1-\beta_k$ and
$\bar{\alpha}_k=\prod_{j=1}^{k}\alpha_j$.
A noisy sample can then be directly obtained from the clean target as
\begin{equation}
\mathbf{x}_k
=
\sqrt{\bar{\alpha}_k}\mathbf{x}_0
+
\sqrt{1-\bar{\alpha}_k}\boldsymbol{\epsilon},
\qquad
\boldsymbol{\epsilon}\sim\mathcal{N}(0,\mathbf{I}).
\end{equation}
As $k$ increases, the spatial structure of the target field is gradually destroyed and the sample approaches Gaussian noise. DDPM learns the reverse process to progressively recover the target field from this noisy representation.

For reservoir state prediction, the reverse process is conditioned on the current state and well-control action,
\begin{equation}
\mathbf{c}_t=(\mathbf{s}_t,\mathbf{a}_t).
\end{equation}
The conditional reverse transition is modeled as
\begin{equation}
p_{\phi}
(\mathbf{x}_{k-1}|\mathbf{x}_k,\mathbf{c}_t)
=
\mathcal{N}
\left(
\boldsymbol{\mu}_{\phi}
(\mathbf{x}_k,k,\mathbf{c}_t),
\boldsymbol{\Sigma}_k
\right),
\end{equation}
where $\phi$ denotes the learnable parameters. Conditioning on $\mathbf{s}_t$ provides the current reservoir configuration, while $\mathbf{a}_t$ introduces the effect of the applied well-control action. The resulting model therefore approximates the controlled transition distribution
$p(\mathbf{s}_{t+1}|\mathbf{s}_t,\mathbf{a}_t)$.

\paragraph{2D Conditional U-Net.}
A 2D conditional U-Net is employed as the denoising network in the DDPM. The encoder progressively extracts multi-scale spatial features from the noisy reservoir field through convolution and downsampling operations, while the decoder reconstructs the spatial resolution through upsampling. Skip connections transfer fine-scale spatial information from the encoder to the decoder, which is important for preserving localized temperature and pressure structures. Residual blocks are further used to facilitate feature propagation and stabilize network training.

At each diffusion step, the U-Net takes the noisy field $\mathbf{x}_k$, the diffusion timestep $k$, and the condition $\mathbf{c}_t$ as inputs. The timestep is encoded through a timestep embedding to represent the current noise level, while the current reservoir state and well-control action guide the denoising process toward the corresponding next reservoir state. The network predicts the noise contained in $\mathbf{x}_k$:
\begin{equation}
\hat{\boldsymbol{\epsilon}}
=
\boldsymbol{\epsilon}_{\phi}
(\mathbf{x}_k,k,\mathbf{c}_t).
\end{equation}

The denoising network is trained using the standard DDPM noise-prediction objective,
\begin{equation}
\mathcal{L}_{\mathrm{diff}}
=
\mathbb{E}
\left[
\left\|
\boldsymbol{\epsilon}
-
\boldsymbol{\epsilon}_{\phi}
(\mathbf{x}_k,k,\mathbf{c}_t)
\right\|_2^2
\right].
\end{equation}
Training samples are constructed from state transitions
$(\mathbf{s}_t,\mathbf{a}_t,\mathbf{s}_{t+1})$
generated by the COMSOL simulator. For the temperature model,
$\mathbf{x}_0=\mathbf{T}_{t+1}$, while for the pressure model,
$\mathbf{x}_0=\mathbf{P}_{t+1}$.

During inference, the reverse DDPM process starts from Gaussian noise and iteratively denoises the sample under the condition $(\mathbf{s}_t,\mathbf{a}_t)$ until the next reservoir field is recovered. The two trained models generate
\begin{equation}
\hat{\mathbf{T}}_{t+1}
=
\mathcal{D}_{T}(\mathbf{s}_t,\mathbf{a}_t),
\qquad
\hat{\mathbf{P}}_{t+1}
=
\mathcal{D}_{P}(\mathbf{s}_t,\mathbf{a}_t),
\end{equation}
which are combined as
$\hat{\mathbf{s}}_{t+1}
=
[\hat{\mathbf{T}}_{t+1},\hat{\mathbf{P}}_{t+1}]$.
The predicted state is then returned to the RL agent for the next control decision, enabling multi-step reservoir rollouts without repeatedly invoking the high-fidelity COMSOL simulator.

\subsection{Reward Surrogate Model}

The diffusion surrogate predicts the next reservoir temperature and pressure fields, but does not directly provide the economic reward required by RL. Therefore, we train a separate reward surrogate to estimate the stage-wise NPV associated with each state transition.

Given the current state $\mathbf{s}_t$, action $\mathbf{a}_t$, and next state $\mathbf{s}_{t+1}$, the reward model predicts
\begin{equation}
\hat{r}_t
=
\mathcal{R}_{\psi}
(\mathbf{s}_t,\mathbf{a}_t,\mathbf{s}_{t+1}),
\end{equation}
where $\mathcal{R}_{\psi}$ denotes the reward surrogate with parameters $\psi$. The model is trained using rewards computed from COMSOL simulations and minimizes the mean squared error
\begin{equation}
\mathcal{L}_{R}
=
\mathbb{E}
\left[
(r_t-\hat{r}_t)^2
\right].
\end{equation}

During policy rollouts, the predicted next state $\hat{\mathbf{s}}_{t+1}$ from the diffusion surrogate is used to estimate the reward,
\begin{equation}
\hat{r}_t
=
\mathcal{R}_{\psi}
(\mathbf{s}_t,\mathbf{a}_t,\hat{\mathbf{s}}_{t+1}).
\end{equation}
Together, the diffusion and reward surrogates provide the next state and reward required for PPO training.

\subsection{PPO-Based Policy Optimization}

With the learned surrogate environment, the well-control problem can be optimized through repeated policy rollouts without directly interacting with the high-fidelity simulator. At control step $t$, the policy $\pi_{\theta}$ takes the current reservoir state $\mathbf{s}_t$ as input and outputs a continuous well-control action $\mathbf{a}_t$. The diffusion surrogate then predicts the next state $\hat{\mathbf{s}}_{t+1}$, while the reward surrogate provides the corresponding reward $\hat{r}_t$. Repeating this interaction over the production horizon generates trajectories for policy optimization.

We employ Proximal Policy Optimization (PPO) to update the control policy. Let $\pi_{\theta_{\mathrm{old}}}$ denote the policy used to collect the current rollout and $\pi_{\theta}$ denote the updated policy. The probability ratio between the two policies is defined as
\begin{equation}
\rho_t(\theta)
=
\frac{\pi_{\theta}(\mathbf{a}_t|\mathbf{s}_t)}
{\pi_{\theta_{\mathrm{old}}}(\mathbf{a}_t|\mathbf{s}_t)}.
\end{equation}
PPO restricts excessive policy updates using the clipped surrogate objective
\begin{equation}
\mathcal{L}_{\mathrm{clip}}
=
\mathbb{E}_t
\left[
\min
\left(
\rho_t(\theta)\hat{A}_t,
\mathrm{clip}
(\rho_t(\theta),1-\epsilon,1+\epsilon)\hat{A}_t
\right)
\right],
\end{equation}
where $\hat{A}_t$ is the estimated advantage and $\epsilon$ controls the clipping range.

A value network $V_{\omega}(\mathbf{s}_t)$ is trained to estimate the expected return of each reservoir state. Its loss is defined as
\begin{equation}
\mathcal{L}_{V}
=
\mathbb{E}_t
\left[
\left(
V_{\omega}(\mathbf{s}_t)-\hat{R}_t
\right)^2
\right],
\end{equation}
where $\hat{R}_t$ denotes the estimated return. An entropy term is further included to encourage exploration of the continuous well-control space. The overall PPO loss is
\begin{equation}
\mathcal{L}_{\mathrm{PPO}}
=
-\mathcal{L}_{\mathrm{clip}}
+
c_v\mathcal{L}_{V}
-
c_e\mathcal{H}(\pi_{\theta}),
\end{equation}
where $c_v$ and $c_e$ control the contributions of the value loss and policy entropy, respectively.

During training, PPO alternates between collecting trajectories from the learned surrogate environment and updating the policy and value networks using the collected transitions. The resulting policy learns to select well-control actions according to the evolving temperature and pressure fields with the objective of maximizing cumulative NPV over the production horizon. After training, the learned policy is evaluated using the original COMSOL simulator.

\begin{algorithm}[t]
\caption{Diffusion-Surrogate PPO for EGS Well-Control Optimization}
\label{alg:diffusion_ppo}
\begin{algorithmic}[1]
\REQUIRE COMSOL dataset $\mathcal{D}$, control horizon $H$
\ENSURE Optimized policy $\pi_\theta$

\STATE Train temperature diffusion model $\mathcal{D}_T$ on $\mathcal{D}$
\STATE Train pressure diffusion model $\mathcal{D}_P$ on $\mathcal{D}$
\STATE Train reward surrogate $\mathcal{R}_\psi$ on $\mathcal{D}$
\STATE Initialize PPO policy $\pi_\theta$ and value network $V_\omega$

\FOR{each PPO iteration}
    \STATE Initialize reservoir state $\mathbf{s}_0$

    \FOR{$t=0,\ldots,H-1$}
        \STATE $\mathbf{a}_t \sim
        \pi_\theta(\cdot|\mathbf{s}_t)$

        \STATE $\hat{\mathbf{T}}_{t+1}
        \leftarrow \mathcal{D}_T(\mathbf{s}_t,\mathbf{a}_t)$

        \STATE $\hat{\mathbf{P}}_{t+1}
        \leftarrow \mathcal{D}_P(\mathbf{s}_t,\mathbf{a}_t)$

        \STATE $\hat{\mathbf{s}}_{t+1}
        \leftarrow
        [\hat{\mathbf{T}}_{t+1},\hat{\mathbf{P}}_{t+1}]$

        \STATE $\hat r_t
        \leftarrow
        \mathcal{R}_\psi(
        \mathbf{s}_t,\mathbf{a}_t,\hat{\mathbf{s}}_{t+1})$

        \STATE Store
        $(\mathbf{s}_t,\mathbf{a}_t,\hat r_t,
        \hat{\mathbf{s}}_{t+1})$

        \STATE $\mathbf{s}_t
        \leftarrow \hat{\mathbf{s}}_{t+1}$
    \ENDFOR

    \STATE Compute advantages using collected trajectories
    \STATE Update $\pi_\theta$ and $V_\omega$ using PPO
\ENDFOR

\STATE Evaluate $\pi_\theta$ using the original COMSOL simulator
\RETURN $\pi_\theta$
\end{algorithmic}
\end{algorithm}

\section{Experiment}

\subsection{Experimental Setup}

We use the same fractured EGS benchmark as Chen et al.~\cite{chen2025multi}. The reservoir contains four injection wells and five production wells. The main physical and operational parameters are listed in Table~\ref{tab:parameter_setting}.

\begin{table}[t]
\centering
\caption{Parameter configurations of the fractured geothermal energy system.}
\label{tab:parameter_setting}
\begin{tabular}{lc}
\hline
Parameter & Value \\
\hline
Initial temperature ($^\circ$C) & 200 \\
Initial pressure (MPa) & 30 \\
Temperature of injected fluid ($^\circ$C) & 20 \\
Bottom-hole pressure of producers (MPa) & 30 \\
Injection rate range of injectors ($\mathrm{m}^3/\mathrm{s}$)
& $[0,200]\times10^{-3}$ \\
Depth (m) & 2,500 \\
Matrix porosity (-) & 0.01 \\
Fracture porosity (-) & 0.1 \\
Matrix permeability ($\mathrm{m}^2$) & $5\times10^{-17}$ \\
Fracture permeability ($\mathrm{m}^2$) & $10^{-9}$ \\
Formation thickness (m) & 40 \\
Matrix heat conductivity (W/(m$\cdot$K)) & 2 \\
Fluid heat conductivity (W/(m$\cdot{}^\circ$C)) & 0.698 \\
Matrix thermal capacity (J/(kg$\cdot{}^\circ$C)) & 850 \\
Fluid thermal capacity (J/(kg$\cdot{}^\circ$C)) & 4,200 \\
\hline
\end{tabular}
\end{table}

The production period is $12{,}000$ days, divided into 20 control stages of 600 days. At each stage, the action consists of the injection rates of four injection wells. The reservoir temperature and pressure fields are represented on $64\times64$ grids and used as the RL state, while the stage-wise NPV is used as the reward.

A total of 512 trajectories are generated using the COMSOL simulator to train the surrogate environment. Each trajectory contains 20 transitions, resulting in 10,240 transitions in total. Among them, 8,700 transitions are used for training and 1,540 for validation. The main hyperparameters of the diffusion surrogate and PPO are given in Table~\ref{tab:hyperparameters}.

\begin{table}[t]
\centering
\caption{Main hyperparameter settings of the diffusion surrogate and PPO.}
\label{tab:hyperparameters}
\begin{tabular}{lll}
\hline
Component & Parameter & Value \\
\hline
Diffusion
& Spatial resolution & $64\times64$ \\
& U-Net channels & $16$, $32$--$64$--$128$--$256$ \\
& Training transitions & 8,700 \\
& Validation transitions & 1,540 \\
& Diffusion steps & 200 \\
& Optimizer & AdamW \\
& Learning rate & $2\times10^{-4}$ \\
& Batch size & 32 \\
& Training steps & 20,000 \\
\hline
PPO
& Action dimension & 4 \\
& Optimizer & Adam \\
& Learning rate & $1\times10^{-5}$ \\
& Discount factor $\gamma$ & 0.99 \\
& GAE parameter $\lambda$ & 0.95 \\
& Clip parameter $\epsilon$ & 0.2 \\
\hline
\end{tabular}
\end{table}

We compare PPO-Surrogate with DE, GPEME, SACOSO, SHPSO, MFSKT, and PPO-COMSOL under the same well-control setting. The surrogate model is used only during policy training, and the final PPO-Surrogate policy is evaluated using the original COMSOL simulator.

\subsection{Overall Performance}

Table~\ref{tab:npv_time} compares the cumulative NPV obtained by different well-control methods. PPO-COMSOL achieves the best overall performance, while PPO-Surrogate obtains the second-best result and consistently outperforms the evolutionary and surrogate-assisted baselines. In particular, PPO-Surrogate remains close to PPO-COMSOL throughout the production horizon, indicating that policy optimization in the learned surrogate environment can recover a high-quality control strategy without repeatedly interacting with the high-fidelity simulator.

\begin{table}[t]
\centering
\caption{Cumulative NPV achieved by different well-control methods at different production stages.}
\label{tab:npv_time}
\begin{tabular}{lccc}
\hline
Method & 3000 d & 6000 d & 12000 d \\
\hline
PPO-COMSOL    & \textbf{1.8201} & \textbf{1.9906} & \textbf{2.1749} \\
PPO-Surrogate & \underline{1.7295} & \underline{1.8776} & \underline{2.0100} \\
MFSKT         & 0.7343 & 1.2947 & 1.9407 \\
SHPSO         & 0.6901 & 1.2676 & 1.8969 \\
SACOSO        & 0.6992 & 1.2170 & 1.8588 \\
GPEME         & 0.5757 & 1.0444 & 1.6725 \\
DE            & 0.4783 & 0.9348 & 1.5234 \\
\hline
\end{tabular}
\end{table}



\begin{figure}
    \centering
    \includegraphics[width=0.95\linewidth]{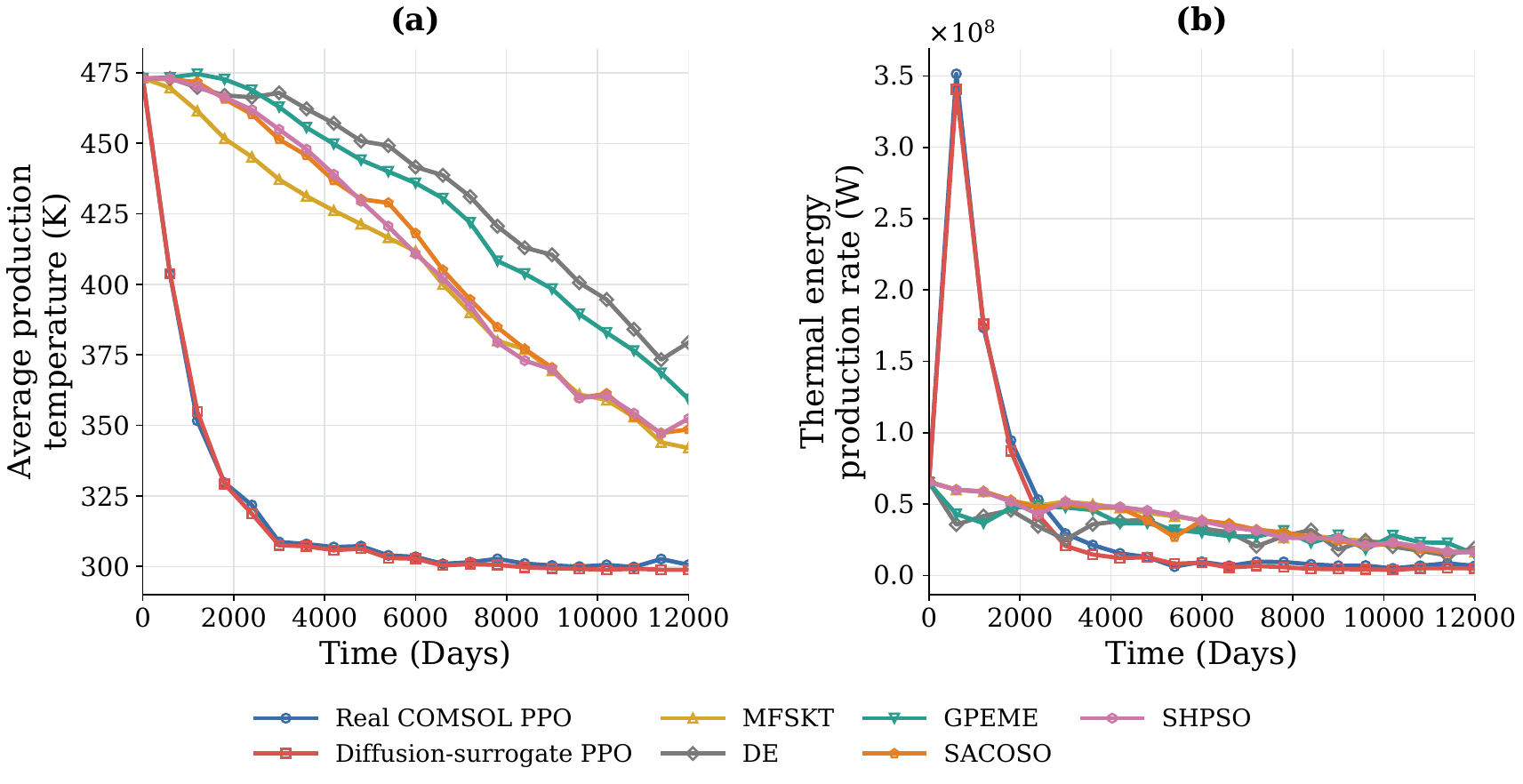}
    \caption{Production performance under different well-control strategies: (a) average production temperature and (b) thermal energy production rate.}
    \label{fig:overall_performance}
\end{figure}

Fig.~\ref{fig:overall_performance} further compares the physical production responses of different methods. The PPO-based strategies exhibit a more aggressive heat-extraction pattern during the early production period, resulting in a higher initial thermal energy production rate and a faster decline in average production temperature. In contrast, the baseline methods maintain a smoother temperature decline but produce lower economic returns. The similar response curves of PPO-COMSOL and PPO-Surrogate further indicate that the surrogate-trained policy captures a control behavior close to that obtained through direct COMSOL-based training.

\subsection{Performance of Surrogate Model}

\begin{figure}[htbp]
    \centering
    \includegraphics[width=0.95\linewidth]{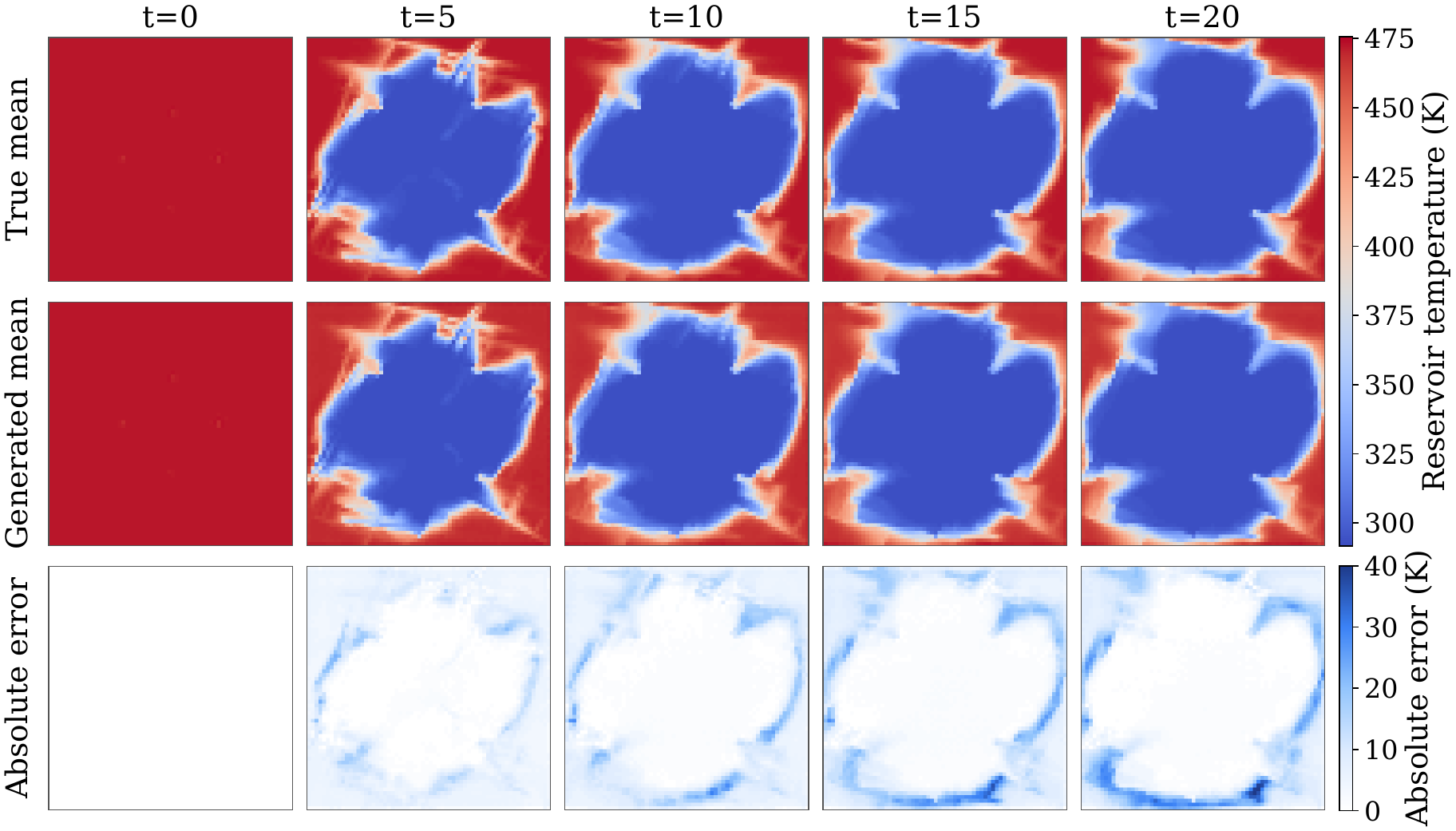}
    \caption{Multi-step temperature-field prediction by the diffusion surrogate.
The rows show the COMSOL reference, surrogate prediction, and absolute
prediction error at different control stages.}
    \label{fig:temperature_prediction}
\end{figure}

\begin{figure}[htbp]
    \centering
    \includegraphics[width=0.95\linewidth]{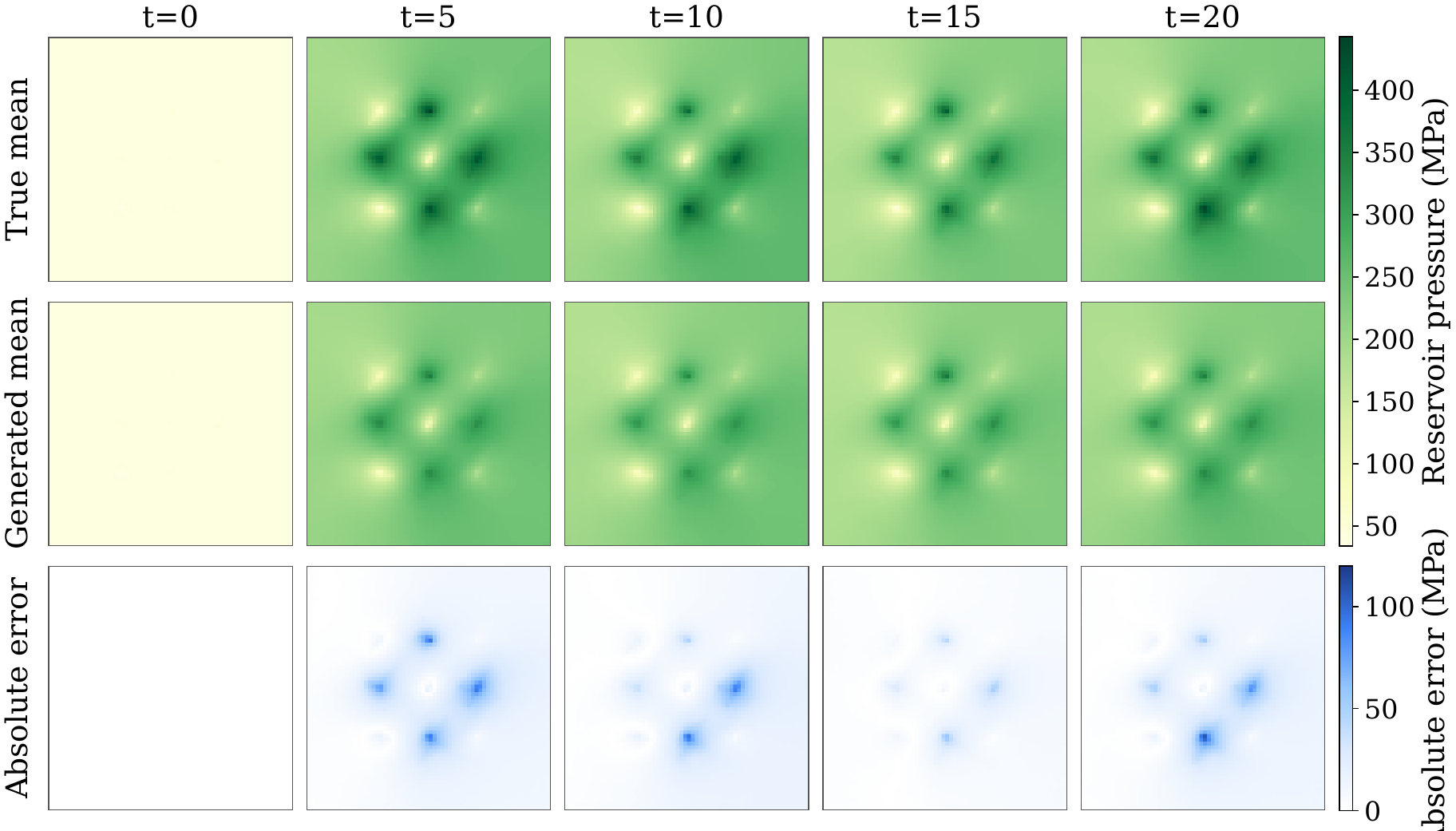}
    \caption{Multi-step pressure-field prediction by the diffusion surrogate.
The rows show the COMSOL reference, surrogate prediction, and absolute
prediction error at different control stages.}
    \label{fig:pressure_prediction}
\end{figure}

We first evaluate whether the learned surrogate can reproduce the reservoir evolution required for long-horizon policy rollouts. Figs.~\ref{fig:temperature_prediction} and~\ref{fig:pressure_prediction} compare the COMSOL states with the states recursively generated by the diffusion surrogate at different control stages. The generated temperature and pressure fields closely reproduce the main spatial patterns of the high-fidelity simulation over the entire production horizon.

For temperature prediction, the surrogate accurately captures the evolution of the thermal front, while the main prediction errors are concentrated around regions with sharp temperature gradients. A similar behavior is observed for pressure prediction, where the generated fields preserve the major pressure distributions and localized responses around the wells. Although prediction errors accumulate during recursive rollout, the overall reservoir structures remain consistent with the COMSOL results up to the final control stage. These results indicate that the diffusion surrogate provides sufficiently stable state transitions for long-horizon PPO training.

\subsection{Well-Control Optimization Performance}

\begin{figure}[htbp]
    \centering
    \includegraphics[width=0.95\linewidth]{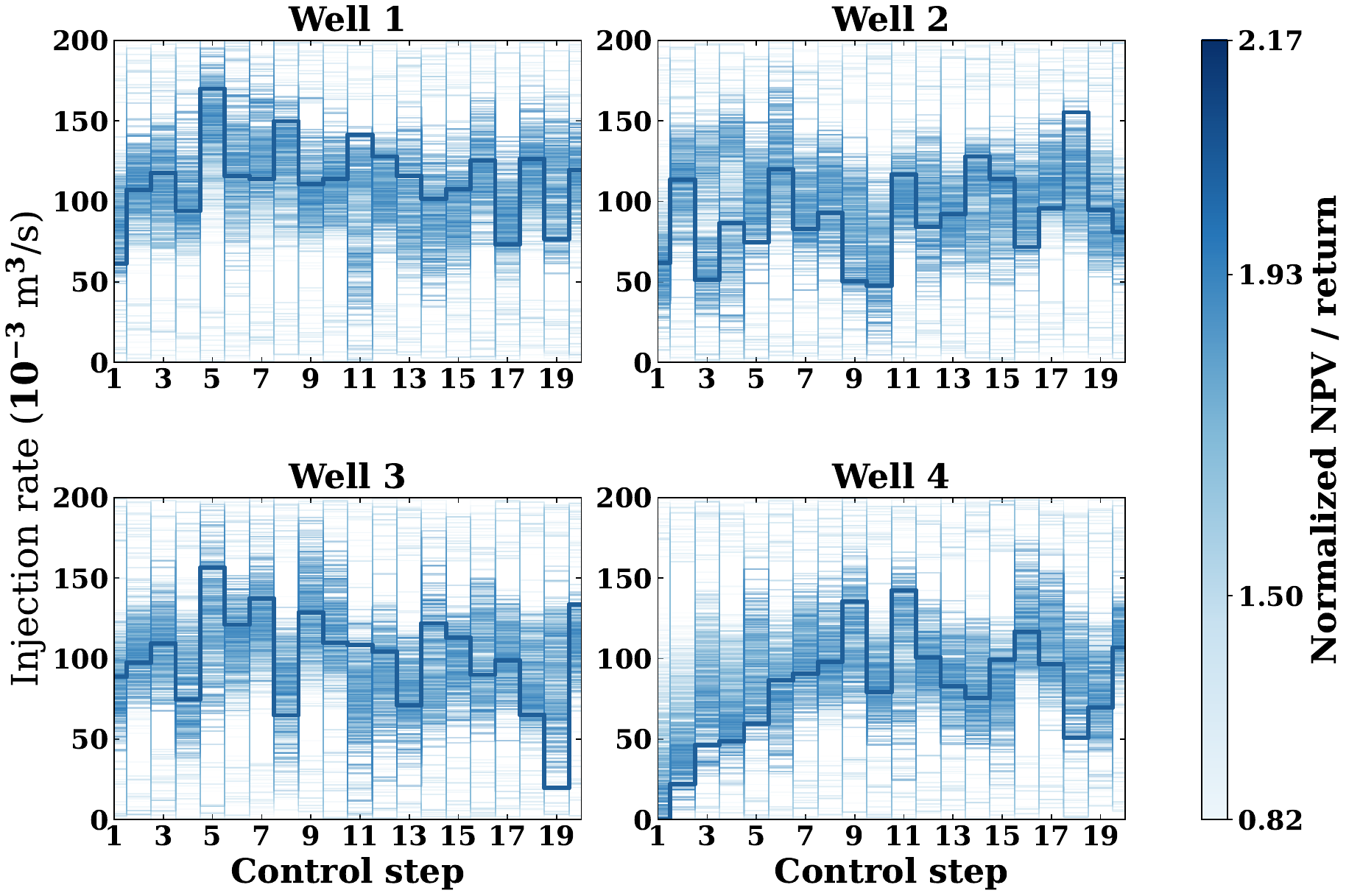}
    \caption{Injection-rate trajectories explored during PPO optimization for
the four injection wells. The color indicates the normalized return of
each trajectory.}
    \label{fig:control_policy}
\end{figure}

To further investigate the learned control strategy, Fig.~\ref{fig:control_policy} visualizes the injection-rate trajectories generated during PPO optimization for the four injection wells. Each trajectory is colored according to its achieved normalized return. The results show that high-return trajectories gradually concentrate within specific injection-rate regions, indicating that PPO learns structured control patterns rather than selecting uniformly distributed actions over the feasible range.

The optimized injection rates also vary across wells and control stages, reflecting the different roles of individual wells in controlling reservoir flow and heat extraction. This state-dependent control behavior allows PPO to adjust the injection strategy according to the evolving temperature and pressure fields instead of optimizing a fixed control sequence independently of the reservoir state.

\subsection{Model Capacity and Data Size Analysis}

\begin{figure}[htbp]
    \centering
    \includegraphics[width=0.98\linewidth]{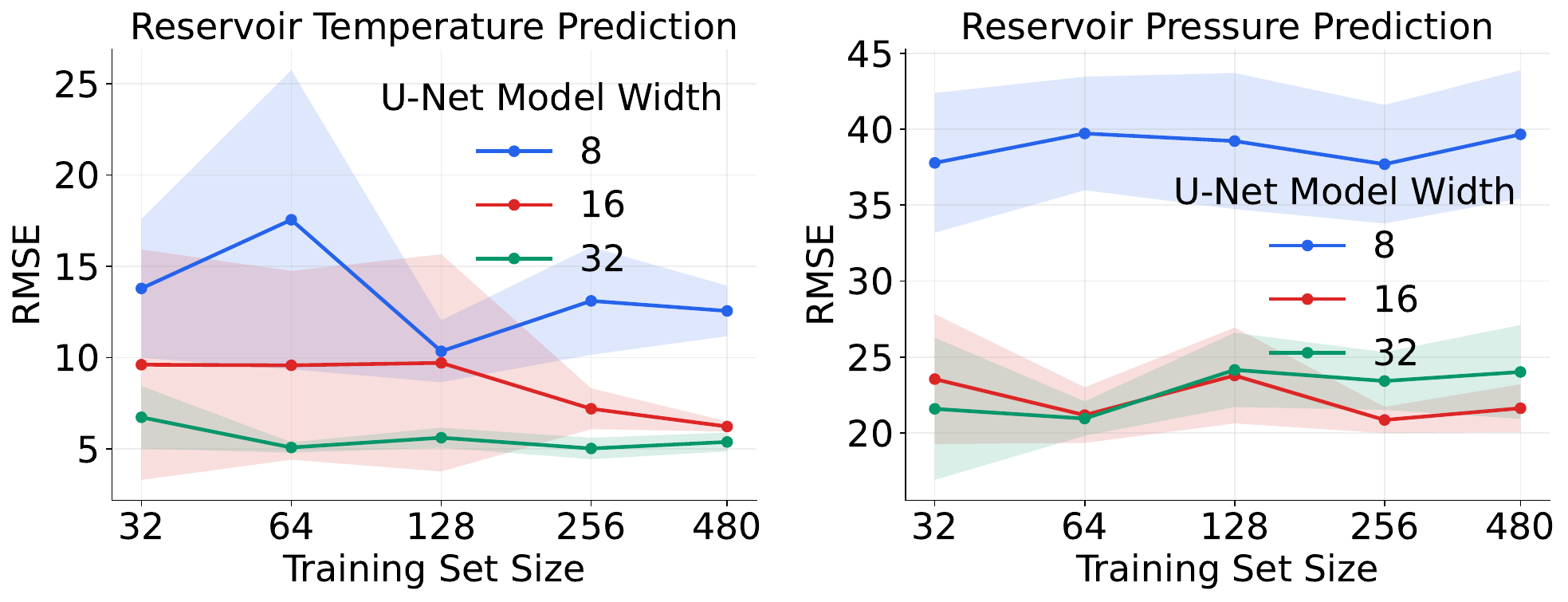}
    \caption{Effects of training-set size and model capacity on autoregressive prediction accuracy for reservoir temperature and pressure.}
    \label{fig:modelcapacitydata}
\end{figure}

We investigate the effects of model capacity and training data size on surrogate prediction accuracy. Model capacity is controlled by the U-Net model width, which determines the number of feature channels in the denoising network, while the training data size refers to the number of rollout trajectories used for surrogate training. We vary the U-Net width among 8, 16, and 32 and the number of training trajectories from 32 to 480. Prediction performance is evaluated using the mean autoregressive test RMSE for reservoir temperature and pressure.

As is shown in Fig.~\ref{fig:modelcapacitydata}, increasing the model width generally improves prediction accuracy. For temperature prediction, the width-32 model achieves the lowest and most stable RMSE, while width-16 and width-32 show comparable performance for pressure prediction. In contrast, increasing the number of training trajectories does not consistently reduce the prediction error. These results suggest that, within the tested range, model capacity has a stronger effect on surrogate accuracy than training data size.

\subsection{Ablation and Cost Analysis}

We first investigate the effect of different reservoir state representations. As shown in Table~\ref{tab:state_ablation}, using both temperature and pressure fields consistently achieves better well-control performance than using either field alone. For PPO-COMSOL, the combined temperature-pressure state achieves a normalized return of 2.17, compared with 2.03 using only temperature and 1.99 using only pressure. The same trend is observed in the surrogate environment, where the combined state achieves 2.01, compared with 1.96 and 1.91 for temperature-only and pressure-only states, respectively. These results suggest that temperature and pressure provide complementary information for state-dependent well-control decisions.

\begin{table}[t]
\centering
\caption{Ablation study on different reservoir state representations.}
\label{tab:state_ablation}
\begin{tabular}{lc}
\hline
Method & Return (Normalized) \\
\hline
PPO-temperature (T)      & 2.03 \\
PPO-pressure (P)         & 1.99 \\
PPO-T-P                  & 2.17 \\
PPO-surrogate-T          & 1.96 \\
PPO-surrogate-P          & 1.91 \\
PPO-surrogate-T-P        & 2.01 \\
\hline
\end{tabular}
\end{table}

\begin{figure}[htbp]
    \centering
    \includegraphics[width=0.95\linewidth]{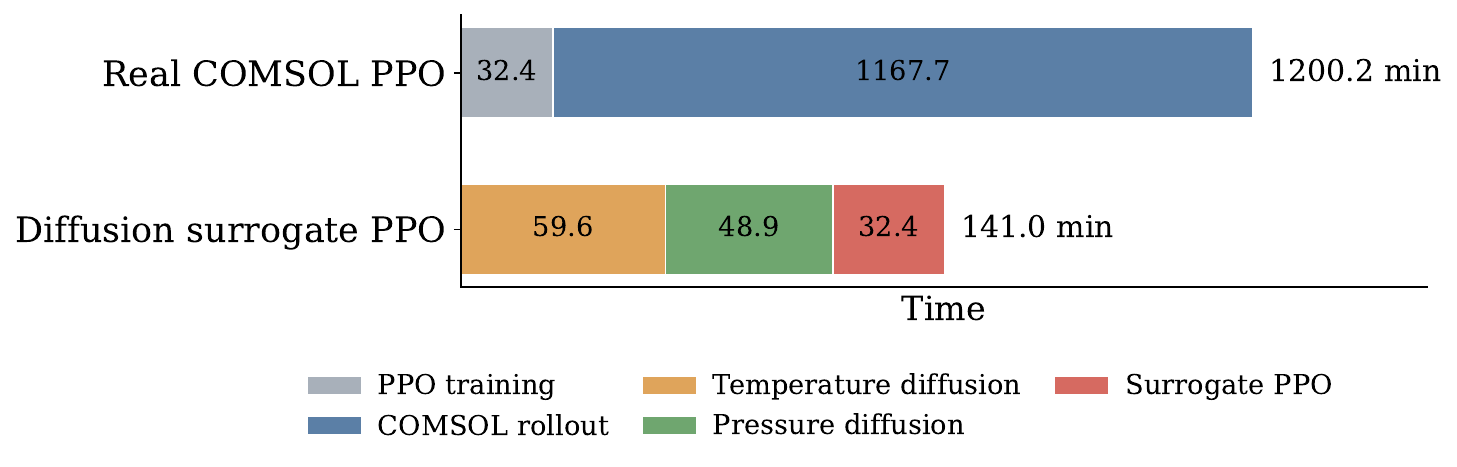}
    \caption{Computational cost of PPO-COMSOL and PPO-Surrogate during
policy optimization.}
    \label{fig:cost}
\end{figure}

Fig.~\ref{fig:cost} compares the computational cost of PPO-COMSOL and PPO-Surrogate. Direct PPO training requires repeated COMSOL rollouts, which dominate the overall computation time. In contrast, PPO-Surrogate replaces these repeated high-fidelity simulations with the learned temperature and pressure diffusion models. The resulting optimization process requires substantially less computation time while maintaining competitive well-control performance. This demonstrates the main computational advantage of using the learned surrogate environment for RL-based EGS optimization.

\begin{figure}
    \centering
    \includegraphics[width=0.95\linewidth]{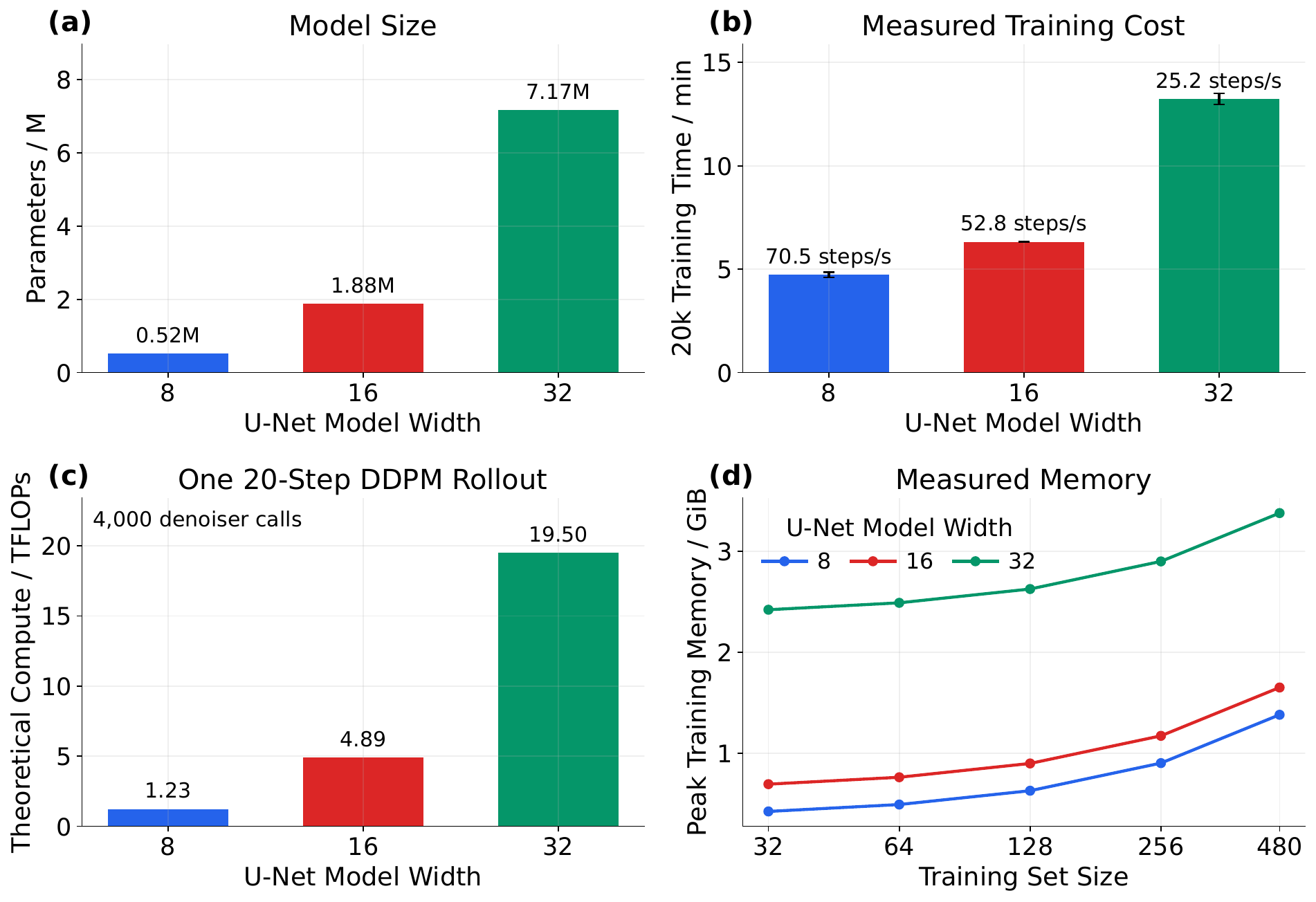}
    \caption{Computational scaling of the diffusion surrogate under different model capacity and training-set sizes. (a) Number of model parameters for different U-Net widths. (b) Measured training time and training throughput for 20k optimization steps. (c) Theoretical computational cost of DDPM rollout. (d) Peak training memory usage under different U-Net widths and training-set sizes.}
    \label{fig:diffusion_model_cost_scaling}
\end{figure}

Fig.~\ref{fig:diffusion_model_cost_scaling} further analyzes the computational cost of the diffusion surrogate under different diffusion model capacities and training-set sizes. Increasing the U-Net width from 8 to 32 substantially increases the number of model parameters, training time, and rollout computation. In particular, the model size increases from 0.52M to 7.17M parameters, while the theoretical cost of a 20-step DDPM rollout increases from 1.23 to 19.50 TFLOPs. The training speed also decreases as the model width increases. In addition, peak memory usage grows with both model width and training-set size.

\subsection{Reward Model Analysis}

\begin{figure}
    \centering
    \includegraphics[width=0.95\linewidth]{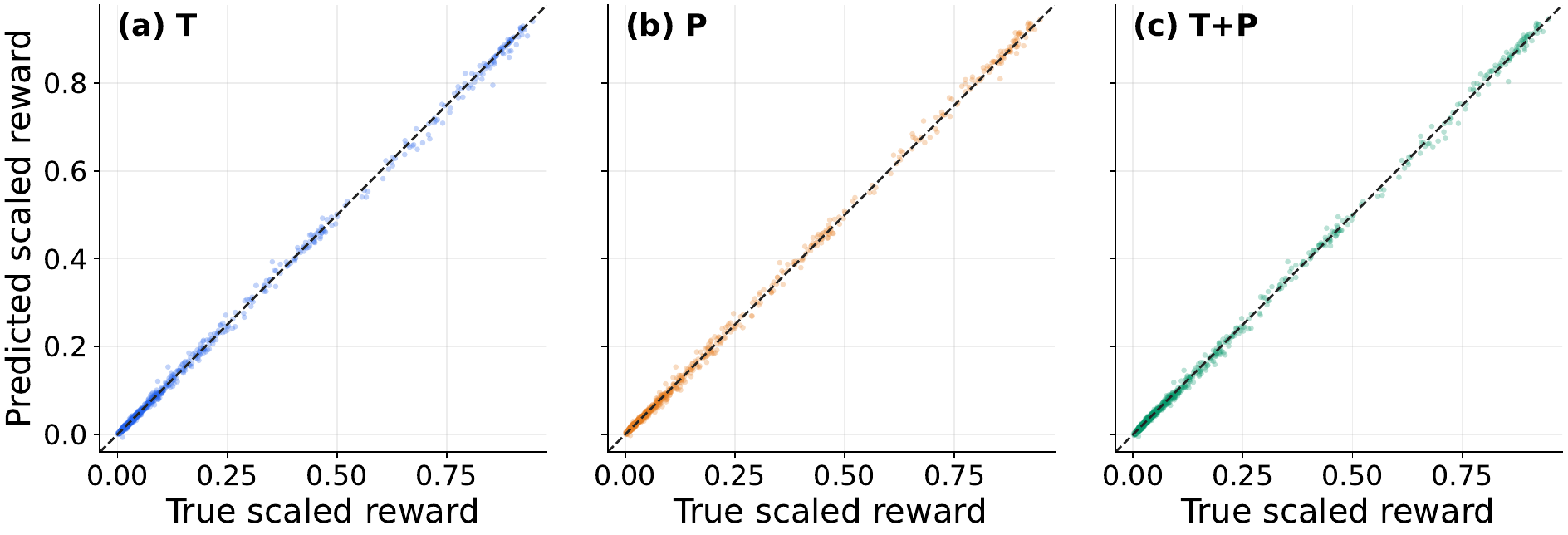}
    \caption{Comparison between true and predicted scaled rewards using different reservoir inputs: (a) temperature (T), (b) pressure (P), and (c) combined temperature and pressure (T+P).}
    \label{fig:reward_predicted_vs_tru}
\end{figure}

We further evaluate the reward surrogate used to provide the economic signal during PPO training. Fig~\ref{fig:reward_predicted_vs_tru} compares the predicted and true scaled rewards using temperature (T), pressure (P), and combined temperature-pressure (T+P) as reservoir inputs. For all three input settings, the predicted rewards closely follow the reference values, indicating that the reward model can accurately reproduce the stage-wise economic returns obtained from the high-fidelity simulations.

\begin{figure}[!t]
    \centering
    \begin{subfigure}[b]{0.48\columnwidth}
        \centering
        \includegraphics[width=\linewidth]{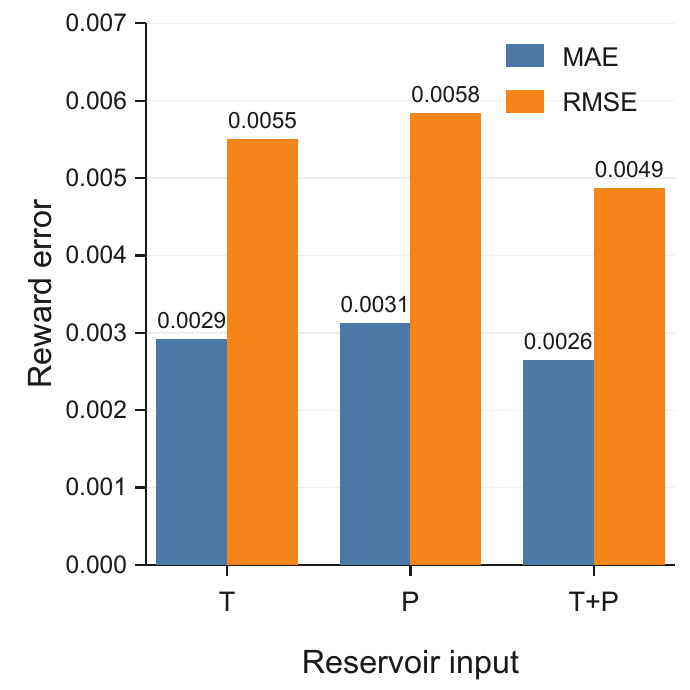}
        \label{fig:a}
    \end{subfigure}
    \hfill
    \begin{subfigure}[b]{0.48\columnwidth}
        \centering
        \includegraphics[width=\linewidth]{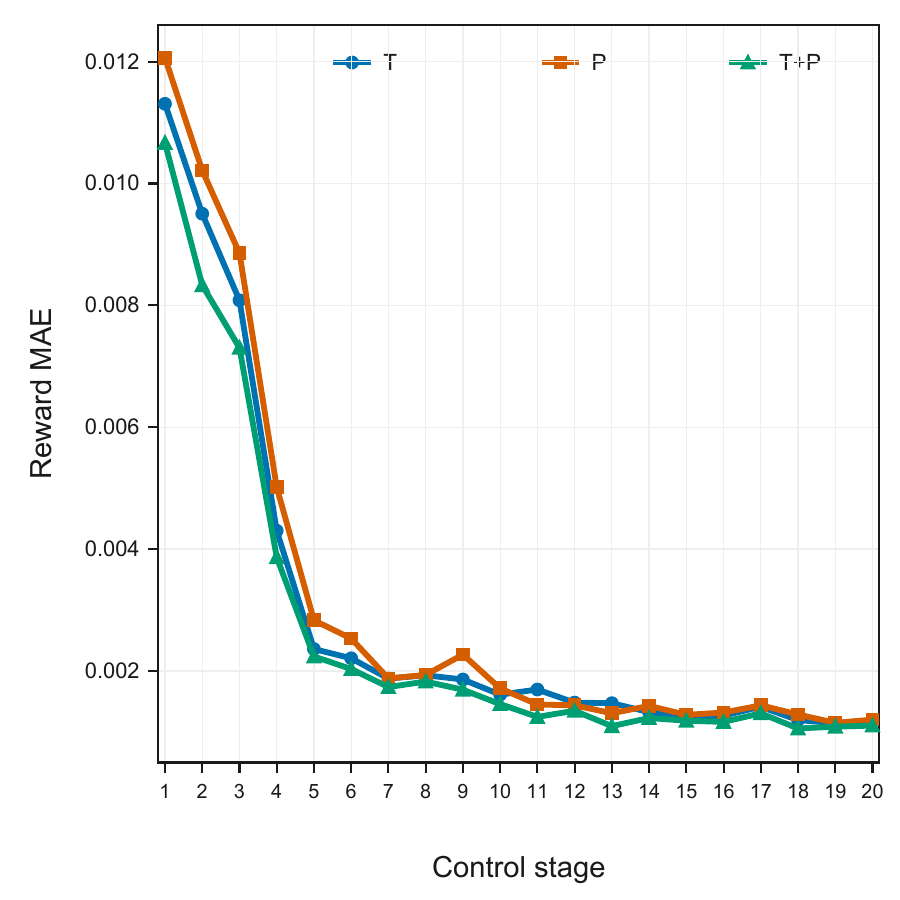}
        \label{fig:b}
    \end{subfigure}
    \caption{Reward prediction performance under different reservoir inputs. (a) Overall MAE and RMSE for models using temperature (T), pressure (P), and combined temperature-pressure (T+P) inputs. (b) Reward MAE at different control stages over the 20-stage production horizon.}
    \label{fig:two_figures}
\end{figure}

Fig~\ref{fig:two_figures} (a) further compares the prediction errors under different reservoir inputs. The T+P model achieves the lowest MAE and RMSE of 0.0026 and 0.0049, respectively, compared with 0.0029 and 0.0055 for the temperature-only model and 0.0031 and 0.0058 for the pressure-only model. These results indicate that temperature and pressure provide complementary information for reward prediction. Fig~\ref{fig:two_figures} (b) shows the reward MAE over the 20 control stages. The prediction errors are relatively larger during the initial stages and decrease rapidly as production proceeds, remaining low throughout the later stages. Overall, the results demonstrate that the reward surrogate provides accurate and stable reward estimates over the production horizon.

\section{Related Work}

Well-control optimization has been extensively studied for improving the economic performance of subsurface energy systems. Conventional approaches mainly formulate the control schedule as a high-dimensional optimization problem and solve it using gradient-based or evolutionary algorithms. Derivative-free methods such as differential evolution (DE) are attractive because they do not require simulator gradients, but typically demand a large number of expensive simulation evaluations. To reduce this cost, surrogate-assisted optimization methods, including GPEME~\cite{liu2013gaussian}, SA-COSO~\cite{sun2017surrogate}, and SHPSO~\cite{yu2018surrogate}, use data-driven models to approximate expensive objective evaluations. More recently, Chen et al.~\cite{chen2025multi} proposed MFSKT for geothermal well-control optimization, combining multi-fidelity surrogate models, knowledge transfer, and active learning to improve search efficiency. Despite their effectiveness, these methods primarily optimize a predefined sequence of well controls rather than learning a state-dependent control policy.

Reinforcement learning (RL) provides an alternative formulation by treating well control as a sequential decision-making problem. Previous studies have applied RL and deep RL to reservoir production optimization and waterflooding control~\cite{hourfar2019reinforcement,ma2019waterflooding}. Zhang et al.~\cite{zhang2022training} formulated life-cycle production optimization as a finite-horizon Markov decision process and employed Soft Actor-Critic (SAC) to learn continuous well-control policies from reservoir pressure and saturation states. Other studies have also investigated PPO-based reservoir control~\cite{nasir2023deep}. These approaches demonstrate the potential of RL to learn feedback policies that directly map evolving reservoir states to control actions. However, direct RL training requires repeated interactions with numerical reservoir simulators, which becomes computationally expensive for high-fidelity hydrothermal models.

Data-driven surrogate models have therefore been increasingly explored to accelerate subsurface simulations. Neural networks, Gaussian processes, and other machine-learning models have been used to approximate reservoir responses and objective functions~\cite{yan2023robust,chen2025multi}. Deep-learning surrogates have also been developed to predict spatial or temporal flow responses in fractured reservoirs~\cite{do2023neural}. Meanwhile, denoising diffusion probabilistic models (DDPMs)~\cite{ho2020denoising} provide a powerful framework for modeling high-dimensional spatial distributions through iterative denoising and have recently shown potential for scientific spatiotemporal forecasting. In contrast to surrogate-assisted optimization that directly approximates the optimization objective, our approach constructs a complete surrogate environment for RL. Conditional diffusion models approximate the controlled evolution of reservoir temperature and pressure fields, while a separate reward surrogate estimates the corresponding economic return. This enables PPO to perform long-horizon policy optimization without repeatedly calling the high-fidelity simulator during training.

\section{Conclusion}

This paper proposes a diffusion-surrogate reinforcement learning framework for EGS well-control optimization. Conditional diffusion models are used to predict the evolution of reservoir temperature and pressure fields, while a separate reward surrogate provides the economic reward required for PPO training. The learned surrogate environment enables state-dependent policy optimization without repeatedly calling the high-fidelity COMSOL simulator during training.

Experiments on a fractured geothermal benchmark show that the proposed PPO-Surrogate method achieves competitive well-control performance compared with existing optimization methods and remains close to PPO trained directly with COMSOL. The results demonstrate that diffusion-based surrogate environments can effectively support long-horizon RL optimization for geothermal well control while reducing dependence on expensive high-fidelity simulations.

\bibliographystyle{IEEEtran}
\bibliography{bigdata}

\end{document}